\makeatletter
\newcommand{\dontusepackage}[2][]{%
  \@namedef{ver@#2.sty}{9999/12/31}%
  \@namedef{opt@#2.sty}{#1}}
\makeatother
\dontusepackage{subfigure}

\documentclass[]{article}

\usepackage{lmodern}
\usepackage{amssymb,amsmath}
\usepackage{ifxetex,ifluatex}
\usepackage[usenames,dvipsnames]{color}
\usepackage{fixltx2e} 
\ifnum 0\ifxetex 1\fi\ifluatex 1\fi=0 
  \usepackage[T1]{fontenc}
  \usepackage[utf8]{inputenc}
\else 
  \ifxetex
    \usepackage{mathspec}
    \usepackage{xltxtra,xunicode}
  \else
    \usepackage{fontspec}
  \fi
  \defaultfontfeatures{Mapping=tex-text,Scale=MatchLowercase}
  
\fi
\IfFileExists{upquote.sty}{\usepackage{upquote}}{}
\IfFileExists{microtype.sty}{%
\usepackage{microtype}
\UseMicrotypeSet[protrusion]{basicmath} 
}{}
\usepackage[margin=1.0in,bottom=1.5in]{geometry}
\usepackage[square,numbers,sort&compress]{natbib}
\usepackage{listings}
\usepackage{graphicx}
\makeatletter
\def\maxwidth{\ifdim\Gin@nat@width>\linewidth\linewidth\else\Gin@nat@width\fi}
\def\maxheight{\ifdim\Gin@nat@height>\textheight\textheight\else\Gin@nat@height\fi}
\makeatother
\setkeys{Gin}{width=\maxwidth,height=\maxheight,keepaspectratio}
\usepackage{caption}
\usepackage{float}
\usepackage{subfig}
\ifxetex
  \usepackage[setpagesize=false, 
              unicode=false, 
              xetex]{hyperref}
\else
  \usepackage[unicode=true]{hyperref}
\fi
\hypersetup{breaklinks=true,
            bookmarks=true,
            pdfauthor={},
            pdftitle={A deep-learning based Bayesian approach to seismic imaging and uncertainty quantification},
            colorlinks=true,
            citecolor=black,
            urlcolor=blue,
            linkcolor=black,
            pdfborder={0 0 0}}
\usepackage[preprint]{neurips_2019}
\usepackage{url}
\usepackage{booktabs}
\usepackage{amsfonts}
\usepackage{nicefrac}
\usepackage{microtype}

\title{A deep-learning based Bayesian approach to seismic imaging and
uncertainty quantification}
\author{Ali Siahkoohi, Gabrio Rizzuti, and Felix J. Herrmann\\School of
Computational Science and Engineering,\\Georgia Institute of
Technology\\\texttt{\{alisk,\phantom{\ }rizzuti.gabrio,\phantom{\ }felix.herrmann\}@gatech.edu}}
\date{}

\begin{document}
\maketitle
\begin{abstract}
Uncertainty quantification is essential when dealing with
ill-conditioned inverse problems due to the inherent nonuniqueness of
the solution. Bayesian approaches allow us to determine how likely an
estimation of the unknown parameters is via formulating the posterior
distribution. Unfortunately, it is often not possible to formulate a
prior distribution that precisely encodes our prior knowledge about the
unknown. Furthermore, adherence to handcrafted priors may greatly bias
the outcome of the Bayesian analysis. To address this issue, we propose
to use the functional form of a randomly initialized convolutional
neural network as an implicit structured prior, which is shown to
promote natural images and excludes images with unnatural noise. In
order to incorporate the model uncertainty into the final estimate, we
sample the posterior distribution using stochastic gradient Langevin
dynamics and perform Bayesian model averaging on the obtained samples.
Our synthetic numerical experiment verifies that deep priors combined
with Bayesian model averaging are able to partially circumvent imaging
artifacts and reduce the risk of overfitting in the presence of extreme
noise. Finally, we present pointwise variance of the estimates as a
measure of uncertainty, which coincides with regions that are more
difficult to image.
\end{abstract}

\section{Introduction}\label{introduction}

\vspace*{-0.25cm}

Seismic imaging involves an inconsistent, ill-conditioned linear inverse
problem due to presence of shadow zones and complex structures in the
subsurface, coherent linearization errors, and noisy measured data. As a
result, there is uncertainty in the recovered unknown parameters.
However, due to the computational complexity of current approaches to
uncertainty quantification in seismic inversion \citep{fang2017uqfip},
most efforts are only focused on computing a maximum a posterior (MAP)
estimate. In this work, we propose using Bayesian-inference, which
provides a principled way of incorporating uncertainty into the
inversion by generating an ensemble of models that each are a solution
to the imaging problem---i.e., sampling the posterior distribution. The
choice of prior in a Bayesian framework is crucial and affects the final
estimate. Conventional methods mostly rely on handcrafted and
unrealistic priors, such as a Gaussian or Laplace distribution prior on
the model parameters in the physical or in a transform domain. However,
handcrafted priors tend to bias the outcome of the inversion, something
we would like to avoid. To address this issue, motivated by earlier
attempts in machine learning and geophysics
\citep{Lempitsky, Cheng_2019_CVPR, wu2019parametric}, we propose to
replace handcrafted priors with an implicit \emph{deep} prior--- i.e.,
reparameterize the unknown with a randomly initialized deep
convolutional neural network (CNN), which is shown to act as a
\emph{structured} prior that promotes \emph{natural} images, but not
unnatural noise. To reduce the risk of overfitting the noise in the
data, we perform Bayesian model averaging by sampling the posterior
using stochastic gradient Langevin dynamics
\citep[SGLD,][]{welling2011bayesian}. Additionally, using the obtained
samples from the posterior, we compute pointwise variance of the
estimates as a measure of uncertainty.

In addition to numerous efforts to incorporate ideas from deep learning
into seismic processing and inversion
\citep{siahkoohi2019transfer, rizzuti2019EAGElis}, in the context of
Bayesian seismic inversion, there has been relatively few attempts
concerning uncertainty qualification. \citet{mosser2018stochastic} first
train a Generative Adversarial Network on synthetic geological
structures. Next, the generator is deployed as an implicit prior in
seismic waveform inversion. Finally, these authors run a variant of SGLD
on the latent variable of the generative model to sample the posterior
and quantify the uncertainty. On the contrary,
\citet{herrmann2019NIPSliwcuc} propose a new formulation in the context
of seismic imaging that does not require a pretrained generative model.
This scheme is based on the Expectation-Maximization method, where they
jointly solve the inverse problem and train a generative model capable
of directly sampling the posterior. The main distinction of their work
is fast posterior sampling since it only requires feed-forward
evaluation of the generative model once the joint inversion and training
is finished.

First, we mathematically formulate the posterior distribution by
introducing the likelihood function and prior distribution involving the
deep prior. Next, we discuss our approach to obtain samples from the
posterior. Finally, we showcase our method using a synthetic example in
the presence of extreme noise.

\section{A Bayesian approach to seismic
imaging}\label{a-bayesian-approach-to-seismic-imaging}

\vspace*{-0.25cm}

In seismic imaging, the goal is to estimate the short-wavelength
structure of the subsurface, denoted by $\delta \mathbf{m}$, given a
smooth background squared-slowness model, $\mathbf{m}_0$, observed data,
$\mathbf{d}_{i}$, and source signatures, $\mathbf{q}_i$, where
$i = 1,2, \cdots , N$ and $N$ is the number of source experiments. Below
we introduce the the likelihood function and prior distribution in our
Bayesian framework.

\subsection{Likelihood function}\label{likelihood-function}

\vspace*{-0.25cm}

We impose a multivariate Gaussian distribution, with a diagonal
covariance matrix, on the noise. If $\mathbf{d}_i$ is $D$ dimensional,
we can write the negative log-likelihood of the observed data as
follows:
\begin{equation}
\begin{aligned}
&\ - \log p_{\text{noise}} \left ( \left \{ \mathbf{d}_{i}, \mathbf{q}_{i} \right \}_{i=1}^N \big{|} \ \delta \mathbf{m} \right )   = -\sum_{i=1}^N  \log p_{\text{noise}} \left ( \mathbf{d}_{i}, \mathbf{q}_{i}  
 \big{|} \ \delta \mathbf{m} \right ) \\ 
&\ = \frac{1}{2 \sigma^2} \sum_{i=1}^N  \|\delta \mathbf{d}_i- \mathbf{J}(\mathbf{m}_0, \mathbf{q}_i) \delta \mathbf{m}\|_2^2 +  \frac{ND}{2}  \log(2 \pi\sigma^2), \quad  \delta \mathbf{d}_i =  \mathbf{d}_i - \mathbf{P}\mathbf{A}(\mathbf{m}_0) ^{-1}\mathbf{q}_i.
\end{aligned}
\label{imaging-likelihood}
\end{equation}
 In these expressions, $p_{\text{noise}}$ is the probability density
function of the noise, $\sigma^2$ is the estimated noise variance,
$\delta \mathbf{d}_i$ the data residual, $\mathbf{J}$ the linearized
Born scattering operator, $\mathbf{A}$ the discretized wave equation,
and $\mathbf{P}$ the restriction operator that restricts the wavefield
to the location of the receivers.

\subsection{Deep prior--- a randomly initialized deep
CNN}\label{deep-prior-a-randomly-initialized-deep-cnn}

\vspace*{-0.25cm}

\begin{figure}
\centering
\subfloat[\label{figure-1c}]{\includegraphics[width=0.500\hsize]{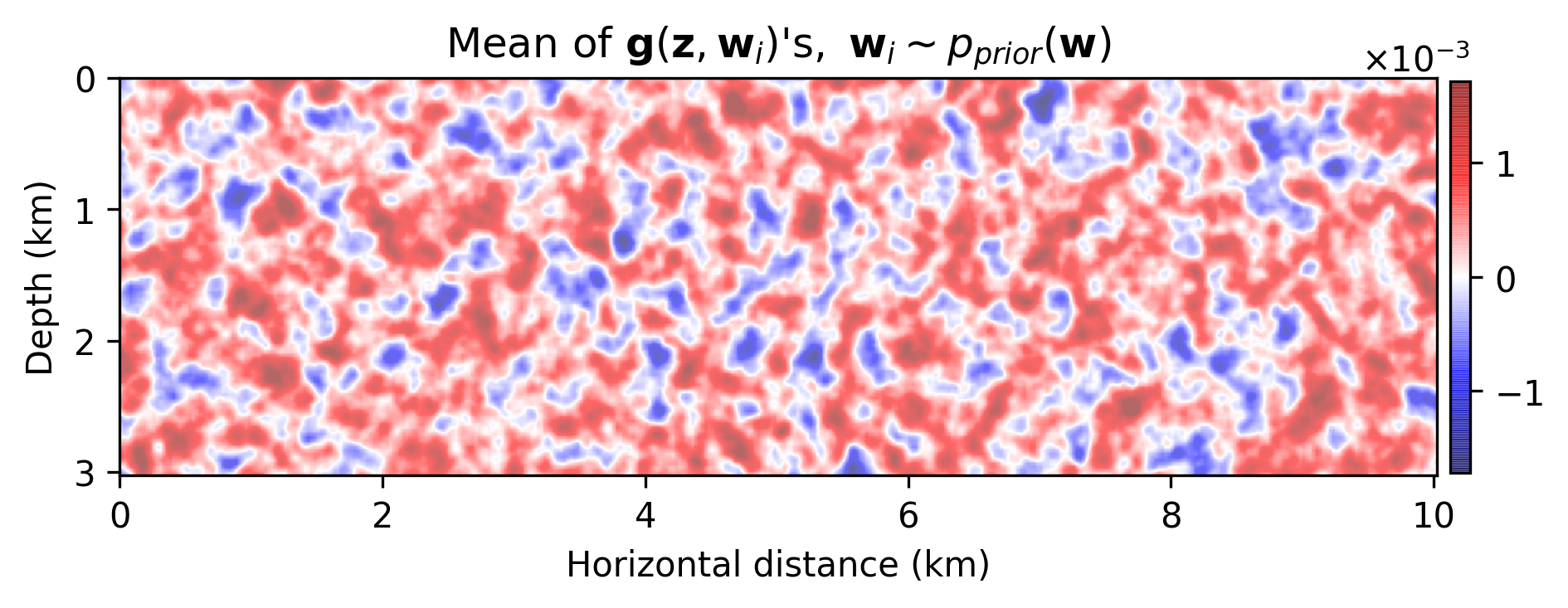}}
\subfloat[\label{figure-1d}]{\includegraphics[width=0.500\hsize]{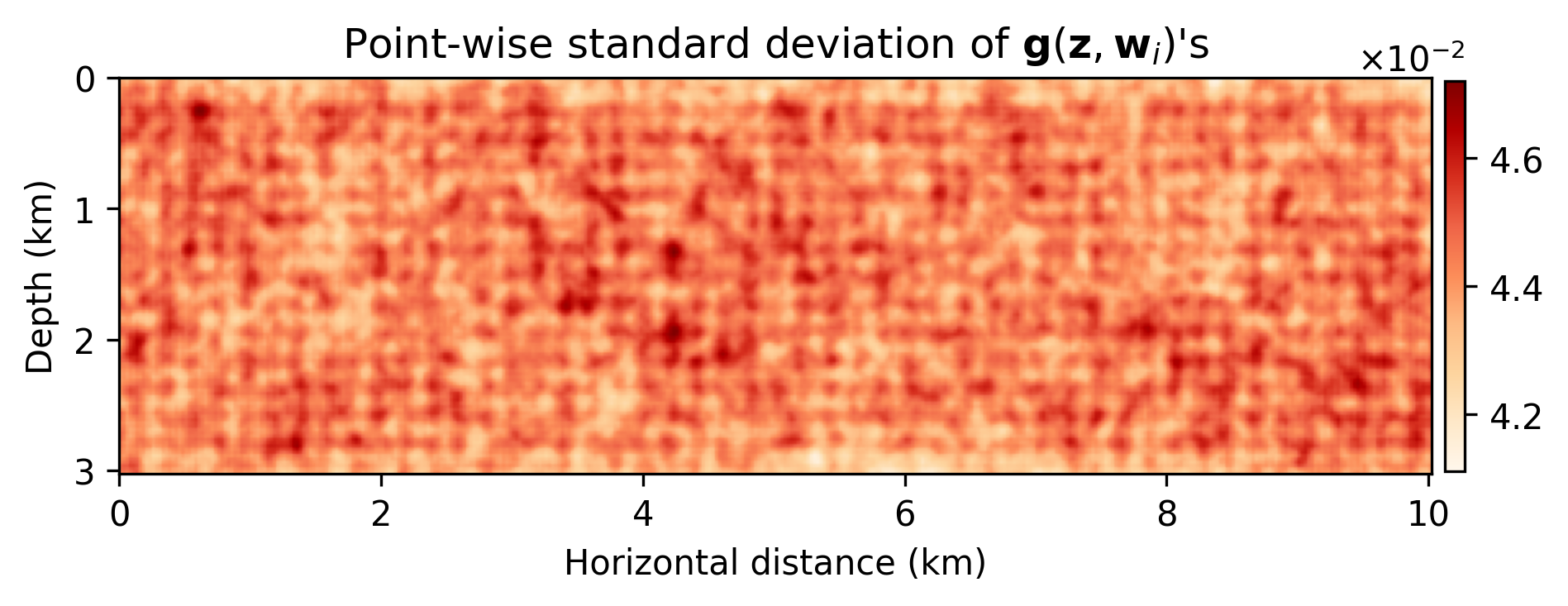}}
\caption{First and second order statistics of the implicit deep prior.
a) The mean of $\mathbf{g} (\mathbf{z}, \mathbf{w})$ over $5000$ samples
$\mathbf{w} \sim p_{\text{prior}}(\mathbf{w})$. b) The pointwise
standard deviation of $\mathbf{g} (\mathbf{z}, \mathbf{w})$ over the
drawn samples.}\label{prior}
\end{figure}

Being a structured prior for natural images, randomly initialized CNNs
are utilized in solving several inverse problems
\citep{Lempitsky, Cheng_2019_CVPR, wu2019parametric}. Motivated by their
success, we propose to reparameterize the unknown model perturbations,
$\delta \mathbf{m}$, by a randomly initialized deep CNN,
$\mathbf{g} (\mathbf{z}, \mathbf{w})$---i.e.,
$\delta \mathbf{m} = \mathbf{g} (\mathbf{z}, \mathbf{w})$, where
$\mathbf{z} \sim \mathrm{N}( \mathbf{0}, \mathbf{I})$ is the fixed input
to the CNN and $\mathbf{w}$ denotes the unknown CNN weights, consisting
of convolutional kernels and biases. We follow \citet{Lempitsky} for the
CNN architecture. We also impose a Gaussian prior on the weights of the
CNN---i.e.,
$\mathbf{w} \sim \mathrm{N}(\mathbf{0}, \frac{1}{\lambda^2}\mathbf{I})$,
where $\lambda$ is a hyperparameter. The seemingly uninformative
Gaussian prior on $\mathbf{w}$ induces a structured prior on the output
space because of the carefully designed functional form of the CNN.
Figure~\ref{prior} demonstrates the empirical first and second order
statistics induced by the deep prior obtained by evaluating the CNN
using $5000$ weights sampled from the prior distribution,
$p_{\text{prior}}(\mathbf{w})$.

Based on the definitions above, we can write the negative log-posterior
as follows:
\begin{equation}
\begin{aligned}
- \log p_{\text{post}} \left ( \mathbf{w} \big{|}  \left \{ \mathbf{d}_{i}, \mathbf{q}_{i} \right \}_{i=1}^N, \mathbf{z} \right )
&\ =  -  \left [ \sum_{i=1}^{N} \log p_{\text{noise}} \left ( \mathbf{d}_{i}, \mathbf{q}_{i}  \big{|} \ \mathbf{w}, \mathbf{z}  \right ) \right ]  - \log p_{\text{prior}} \left ( \mathbf{w} \right ) \quad + \underbrace {\text{const}}_{\text{independent of } \mathbf{w}} \\
&\ = \frac{1}{2 \sigma^2} \sum_{i=1}^N  \|\delta \mathbf{d}_i- \mathbf{J}(\mathbf{m}_0, \mathbf{q}_i) \mathbf{g} (\mathbf{z}, \mathbf{w}) \|_2^2 + \frac{\lambda^2}{2} \| \mathbf{w} \|_2^2 \quad + \quad  \text{const,} \\
\end{aligned}
\label{imaging-obj}
\end{equation}
 where $p_{\text{post}}$ is the posterior distribution. In the next
section, we present how we reap information on the posterior
distribution,
$p_{\text{post}} ( \mathbf{w} \big{|} \left \{ \mathbf{d}_{i}, \mathbf{q}_{i} \right \}_{i=1}^N, \mathbf{z} )$.

\subsection{Sampling the posterior--- stochastic gradient Langevin
dynamics}\label{sampling-the-posterior-stochastic-gradient-langevin-dynamics}

\vspace*{-0.25cm}

The minimizer of the negative log-posterior (Equation~\ref{imaging-obj})
with respect to $\mathbf{w}$ is the MAP estimate. Sampling the posterior
distribution, instead of simply computing the MAP estimate, allows us to
incorporate the model uncertainty into the final estimate
(Equation~\ref{integration}). While Bayesian inference in deep CNNs is
generally intractable, a popular approach to sample the posterior is
SGLD \citep{welling2011bayesian}. As described in Equation~\ref{sgld},
SGLD is a Markov chain Monte Carlo sampler obtained by injecting
Gaussian noise to stochastic gradient descent updates---i.e.,
\begin{equation}
\begin{aligned}
&\ \mathbf{w}_{k+1} = \mathbf{w}_{k} - \frac{\epsilon}{2} \nabla_{\mathbf{w}} J^{(i)}(\mathbf{w}_{k}) + \boldsymbol{\eta}_k, \quad \boldsymbol{\eta}_k \sim \mathrm{N}( \mathbf{0}, \epsilon\mathbf{I}), \\
\end{aligned}
\label{sgld}
\end{equation}
 where
$J^{(i)} (\mathbf{w}) = \frac{N}{2 \sigma^2} \|\delta \mathbf{d}_i- \mathbf{J}(\mathbf{m}_0, \mathbf{q}_i) \mathbf{g} (\mathbf{z}, \mathbf{w}) \|_2^2 + \frac{\lambda^2}{2} \| \mathbf{w} \|_2^2$
approximates the negative log-posterior (Equation~\ref{imaging-obj}) by
using the $i^{\text{th}}$ element in the sum. We integrate Devito's
\citep{devito-api} linearized Born scattering operator into the PyTorch
\citep{NEURIPS2019_9015} deep learning library, thus allowing us to
compute the gradients required in Equation~\ref{sgld} with automatic
differentiation. Our implementation can be found on
\href{https://github.com/alisiahkoohi/seismic-imaging-with-SGLD}{GitHub}.
Finally, we compute the final estimation by Bayesian model averaging as
follows:
\begin{equation}
\begin{aligned}
\delta \widehat { \mathbf{m}} &\ = \mathbb{E}_{\mathbf{w} \sim p_{\text{post}} ( \mathbf{w} \normalsize{|}  \left \{ \mathbf{d}_{i}, \mathbf{q}_{i} \right \}_{i=1}^N, \mathbf{z} )} \left [ \mathbf{g}( \mathbf{z}, \mathbf{w}) \right ] = \int  p_{\text{post}} ( \mathbf{w} \normalsize{|}  \left \{ \mathbf{d}_{i}, \mathbf{q}_{i} \right \}_{i=1}^N, \mathbf{z} ) \mathbf{g}( \mathbf{z}, \mathbf{w}) d\mathbf{w} \\
&\ \simeq \frac{1}{T}\sum_{j=1}^{T} \mathbf{g}( \mathbf{z}, \widehat{\mathbf{w}}_j), \quad \widehat{\mathbf{w}}_j \sim p_{\text{post}} ( \mathbf{w} \normalsize{|}  \left \{ \mathbf{d}_{i}, \mathbf{q}_{i} \right \}_{i=1}^N, \mathbf{z} ), \ j=1, \ldots, T\, ,
\end{aligned}
\label{integration}
\end{equation}
 where $T$ is the number of samples from the posterior distribution and
$\widehat{\mathbf{w}}_j$'s denote the samples.

\section{Numerical experiment}\label{numerical-experiment}

\vspace*{-0.25cm}

We apply our framework to a synthetic dataset simulated on the $2$D
Overthrust model by solving the acoustic wave equation. Our dataset
includes $369$ shot records with $369$ receivers separated by $27$
$\mathrm{m}$, $2$ seconds recording time, and a source wavelet with $8$
$\mathrm{Hz}$ central frequency. We add Gaussian noise drawn from
$\mathrm{N}(\mathbf{0}, 2\mathbf{I})$ to shot records. Taking into the
account the linearization error, the signal-to-noise ratio of the
observed data is $-11.37$ $\mathrm{dB}$. We generate simultaneous source
experiments by mixing the shot records according to $369$ normally
distributed source encodings. By conducting extensive parameter tuning,
we set $\lambda = 170$, $\epsilon = 0.002$, and we run $10000$ total
SGLD iterations. After the first $3000$ burn-in iterations, we select
every $50^{\text{th}}$ update of SGLD as a samples from posterior to
reduce correlation among samples. We also set $\sigma^2 = 2.24$, which
is the summation of the measurement noise variance and a direct
estimation of linearization error variance using the ground-truth
perturbation model.

\begin{figure}
\centering
\subfloat[\label{results-a}]{\includegraphics[width=0.500\hsize]{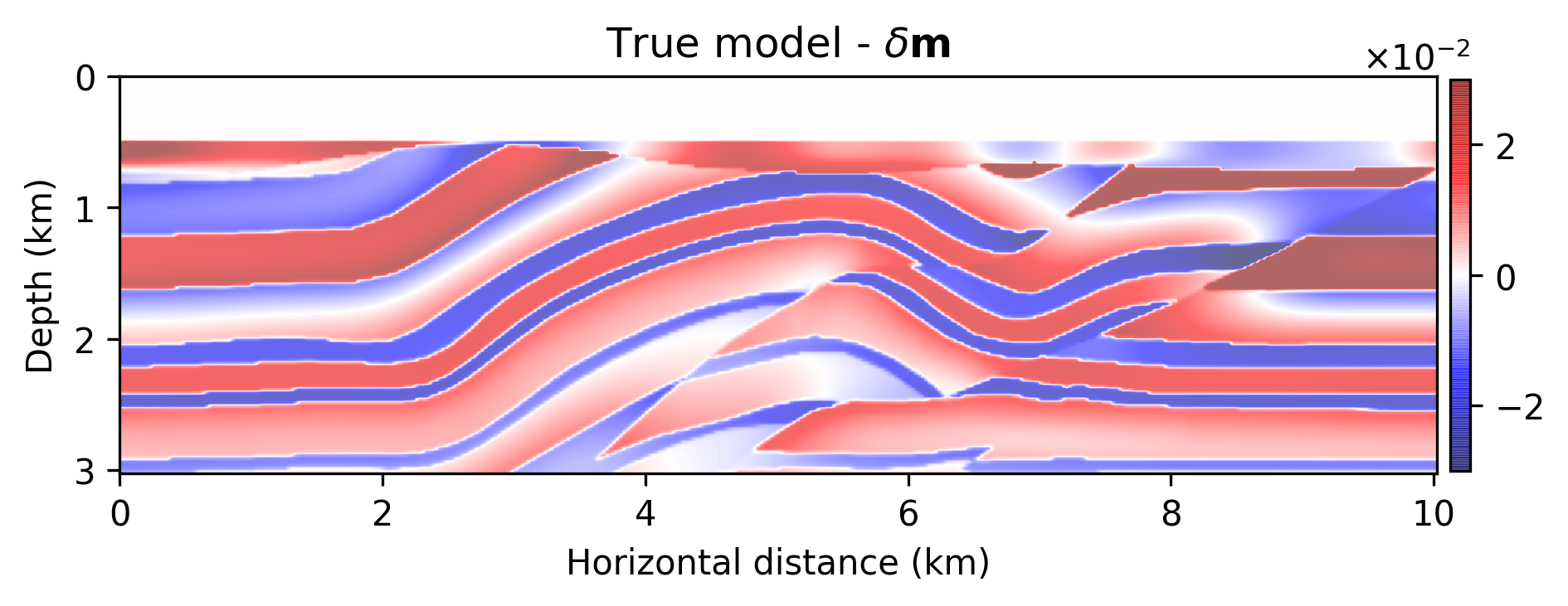}}
\subfloat[\label{results-c}]{\includegraphics[width=0.500\hsize]{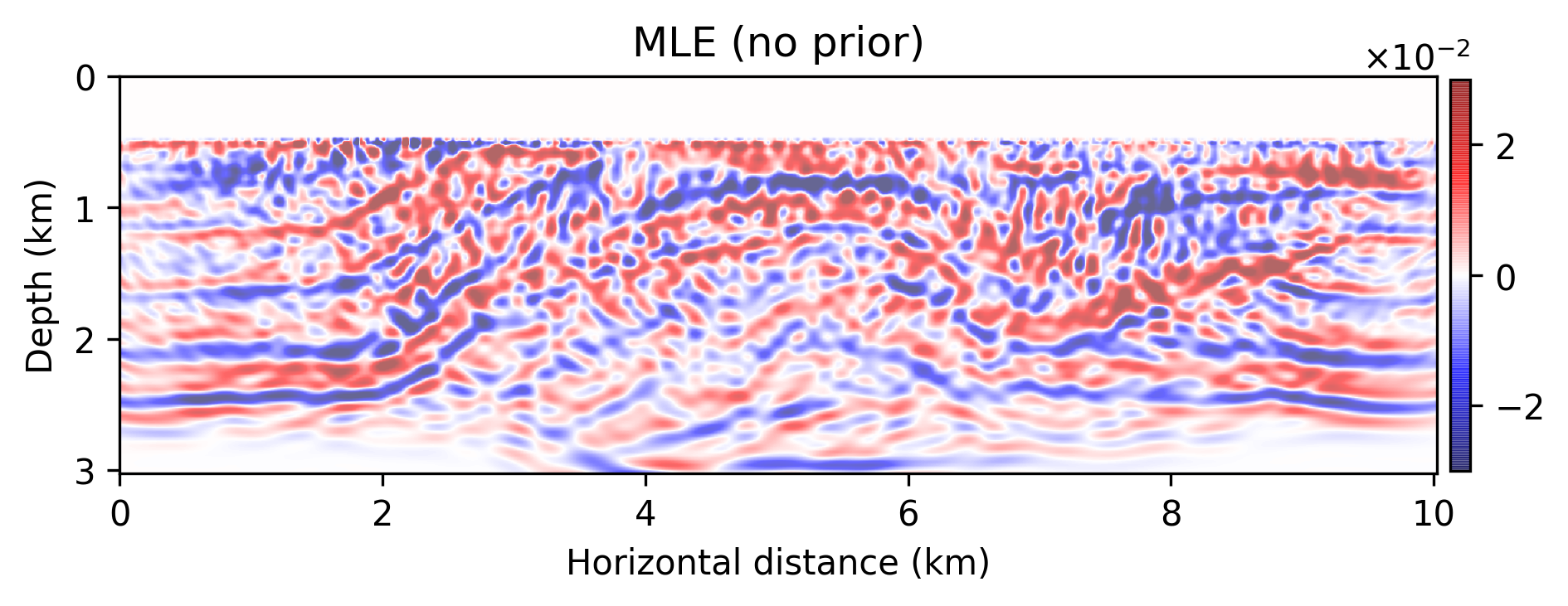}}
\\
\subfloat[\label{results-d}]{\includegraphics[width=0.500\hsize]{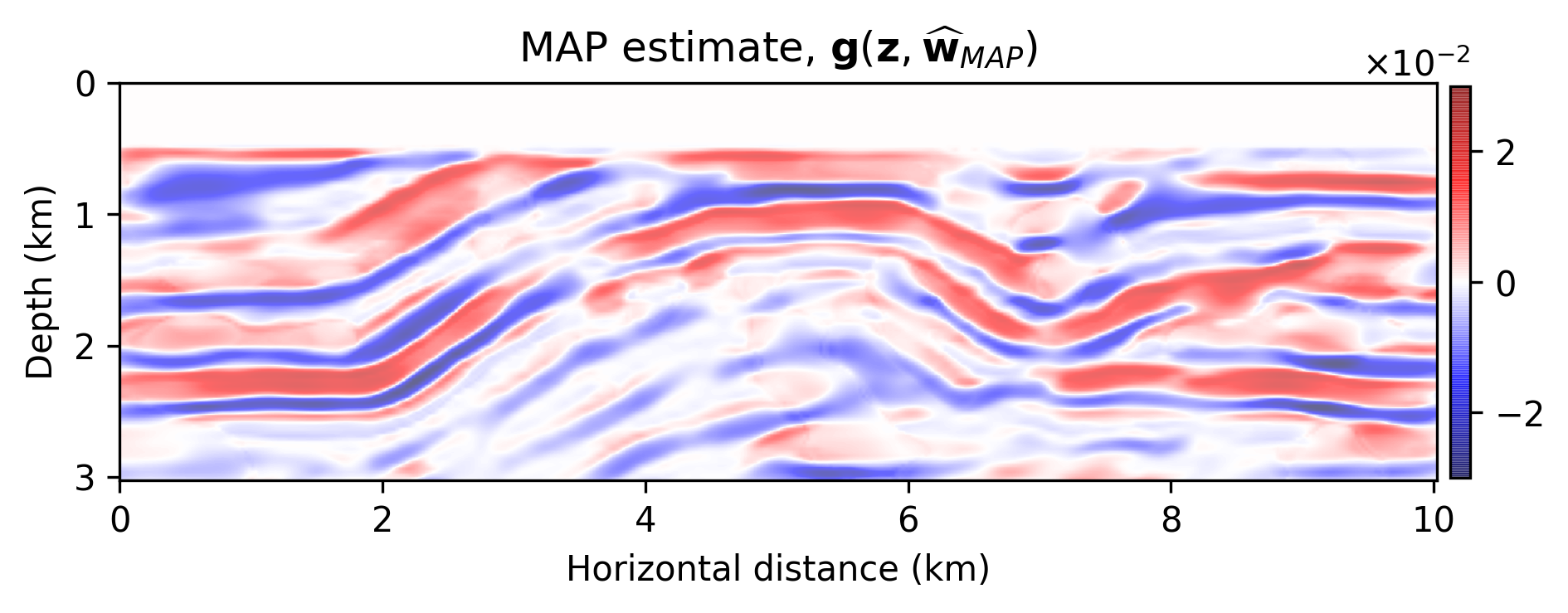}}
\subfloat[\label{results-b}]{\includegraphics[width=0.500\hsize]{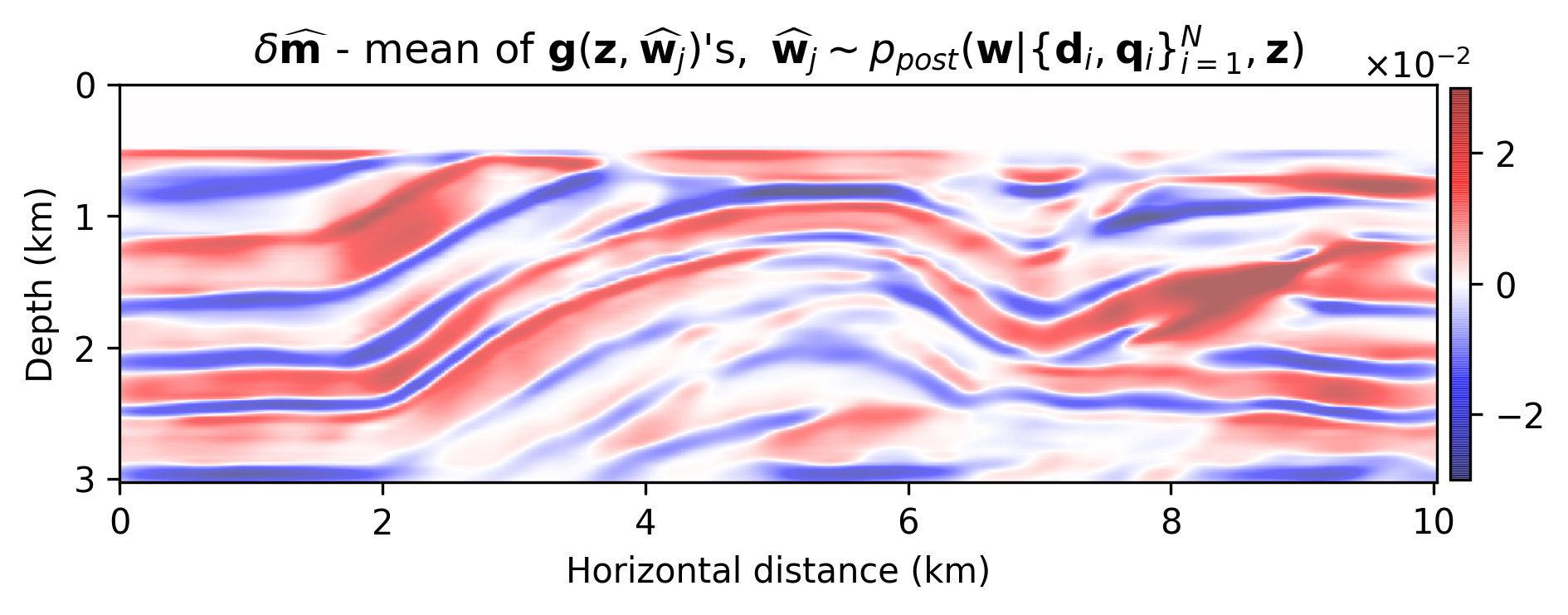}}
\\
\subfloat[\label{results-e}]{\includegraphics[width=0.500\hsize]{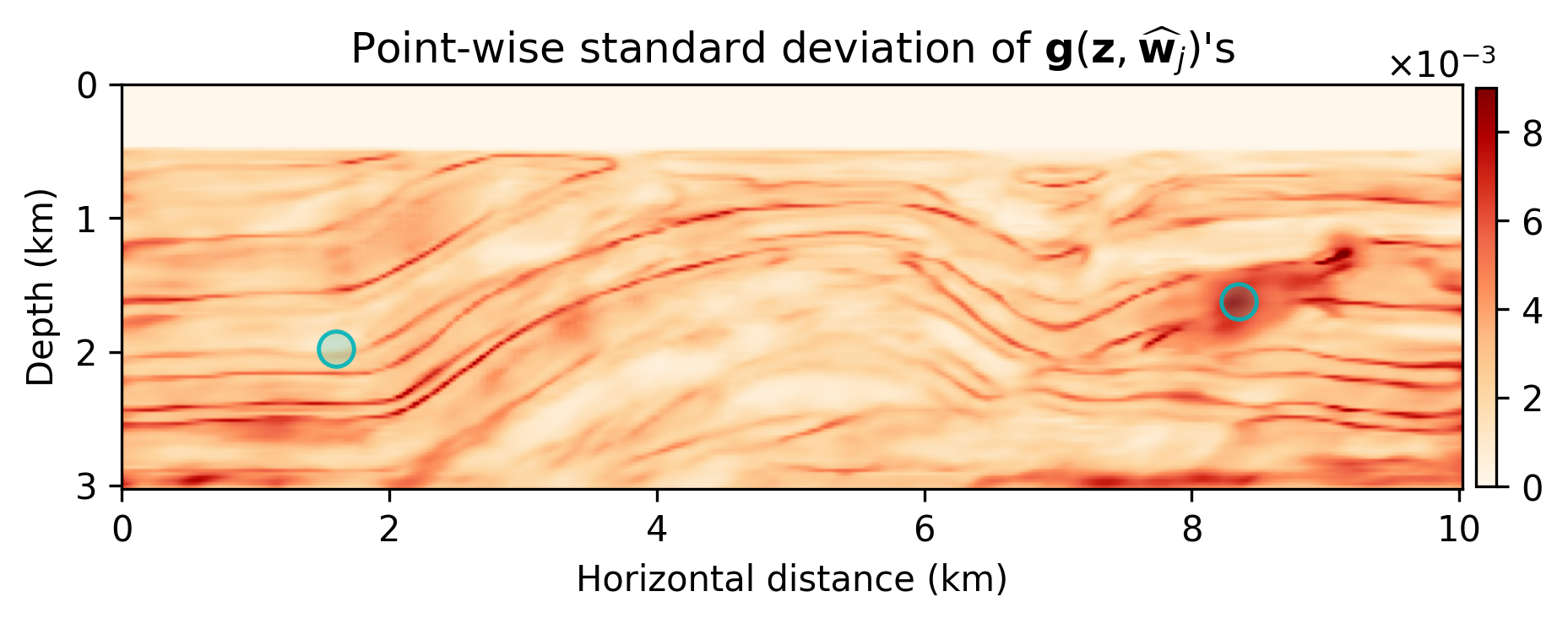}}
\subfloat[\label{results-f}]{\includegraphics[width=0.500\hsize]{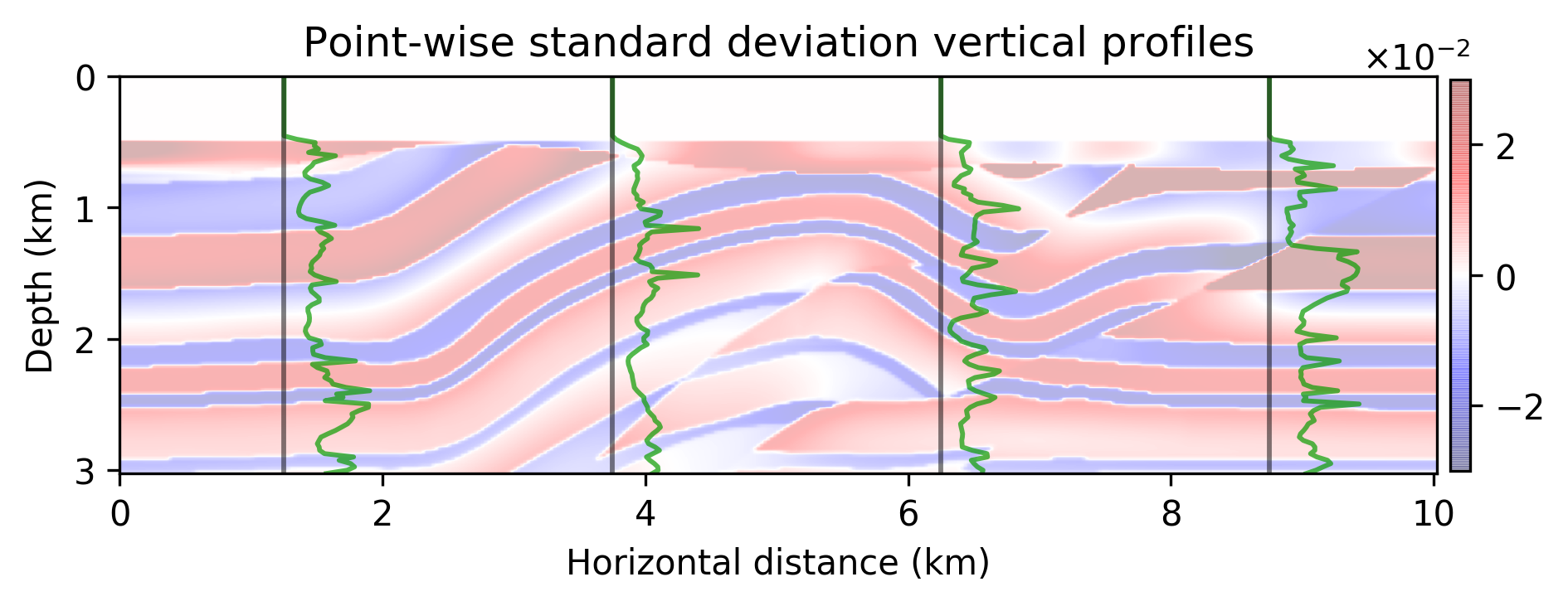}}
\\
\subfloat[\label{results-g}]{\includegraphics[width=0.490\hsize]{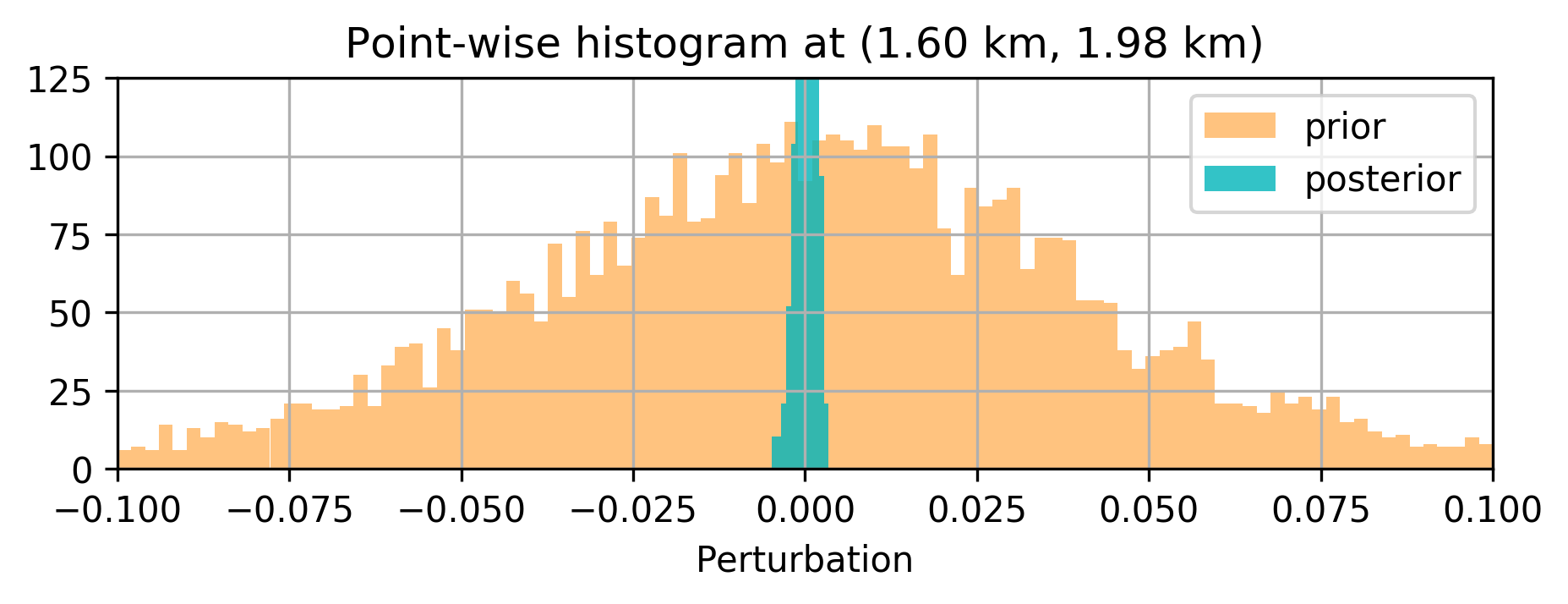}}
\subfloat[\label{results-h}]{\includegraphics[width=0.490\hsize]{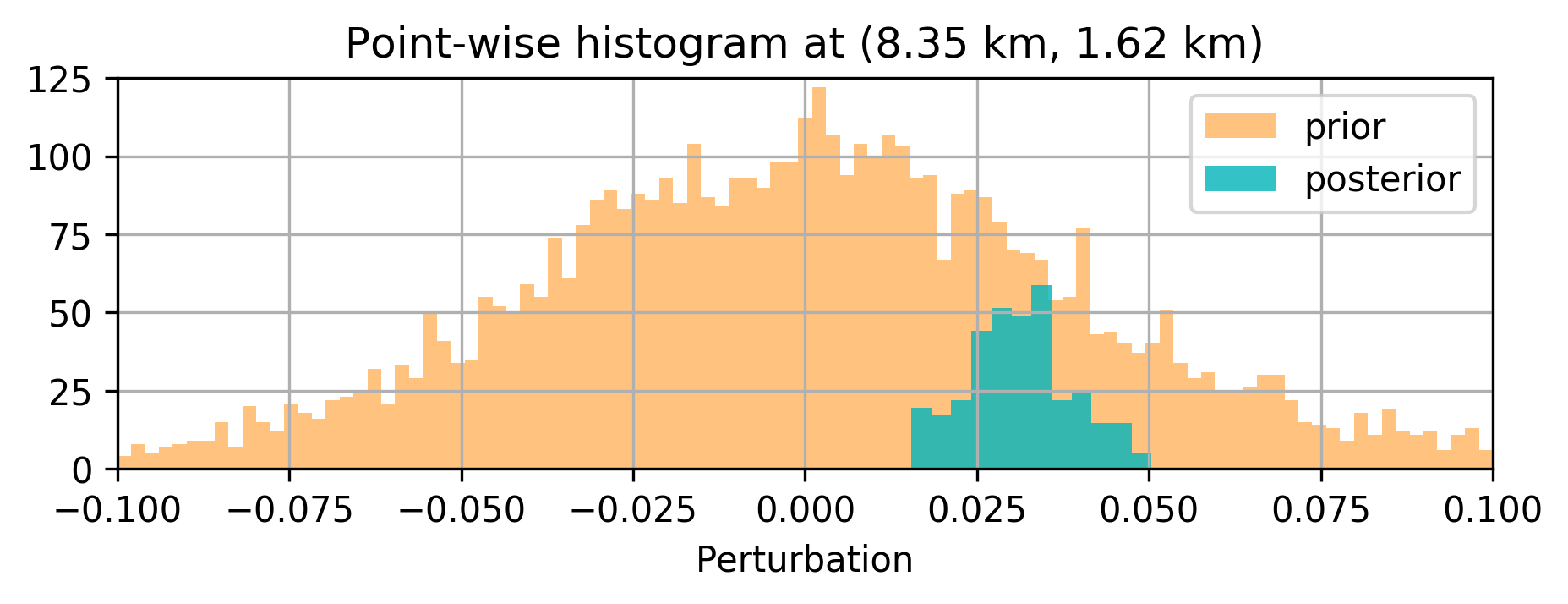}}
\caption{Imaging and uncertainty quantification. a) True model. b) The
MLE ---i.e., minimizer of Equation~\ref{imaging-likelihood} with respect
to $\delta \mathbf{m}$. c) The MAP estimate---i.e., minimizer of
Equation~\ref{imaging-obj}. d) The final estimate,
$\delta \widehat { \mathbf{m}}$. e) The pointwise standard deviation
among samples drawn from the posterior. f) Standard-deviation profiles.
g, h) Point-wise histogram plots at two points in the model indicated by
blue dots in Figure~\ref{results-e}.}\label{results}
\end{figure}

The results are included in Figure~\ref{results}. Figure~\ref{results-c}
is the maximum-likelihood estimate (MLE)--- i.e., using no prior
distribution, obtained by minimizing Equation~\ref{imaging-likelihood}
with respect to $\delta \mathbf{m}$ with stochastic optimization and
early stopping to prevent overfitting the noise, Figure~\ref{results-d}
offers a comparison to the MAP estimate, computed by minimizing
Equation~\ref{imaging-obj}, Figure~\ref{results-b} indicates the
estimation via the proposed method, $\delta \widehat { \mathbf{m}}$,
Figure~\ref{results-e} is the pointwise standard deviation among the
samples from the posterior, and Figure~\ref{results-f} overlays the
vertical profiles of the pointwise standard deviation onto the
ground-truth model. The black lines indicate the horizontal locations
for which we plot the standard-deviation profiles (green lines).
Figures~\ref{results-g} and~\ref{results-h} show pointwise histograms of
prior and posterior distributions at points indicated by blue circles in
Figure~\ref{results-e}. We make the following observations. The prior
induced by the architecture of the CNN, without any other prior
knowledge, has been successful in generating a reasonable image
(Figure~\ref{results-d}) compared to the MLE. The final estimate,
$\delta \widehat { \mathbf{m}}$, is smoother than the MAP estimate and
contains fewer imaging artifacts. Figure~\ref{results-e} indicates that
we have the most uncertainty at the location of the reflectors, and it
gets slightly larger by depth, close to boundaries and fault zone, which
are more difficult to image. Finally, Figures~\ref{results-g}
and~\ref{results-h} indicate sharpening of the histogram after Bayesian
inference.

\section{Discussion and conclusions}\label{discussion-and-conclusions}

\vspace*{-0.25cm}

We introduced a Bayesian framework for seismic imaging that instead of
adhering to handcrafted priors utilizes a structured prior induced by a
carefully designed convolutional neural network. We demonstrated that
our approach is capable of sampling the posterior distribution by
running stochastic gradient Langevin dynamics, albeit being expensive,
similar to most Markov chain Monte Carlo sampling based approaches. Not
withstanding these costs, our formulation is to our knowledge an early
attempt to quantify the uncertainty of a convolutional neural
network-regularized linear inversion jointly capturing uncertainty in
the imaging and the reparametrization with a convolutional neural
network. As verified by our numerical experiment, the utilized deep
prior was partially able to circumvent the imaging artifacts caused by
strong measurement noise and linearization errors. By sampling the
posterior and performing Bayesian model averaging, we were able to
decrease the artifacts. Finally, the pointwise standard variation plot
pointed out more uncertainty at the location of the reflectors, faults,
edges, and deeper parts of the model, which coincide with regions that
are more difficult to image. As a future direction, we propose to avoid
Markov chain Monte Carlo samplers due to their computational complexity.

\bibliography{UQwDIP}

\end{document}